%% file: template.tex
\documentclass{article}

\usepackage{arxiv}

\usepackage[utf8]{inputenc} % allow utf-8 input
\usepackage[T1]{fontenc}    % use 8-bit T1 fonts
\usepackage{hyperref}       % hyperlinks
\usepackage{url}            % simple URL typesetting
\usepackage{booktabs}       % professional-quality tables
\usepackage{amsfonts}       % blackboard math symbols
\usepackage{nicefrac}       % compact symbols for 1/2, etc.
\usepackage{microtype}      % microtypography
\usepackage{lipsum}
\usepackage{graphicx}
\usepackage{orcidlink}
\usepackage{array}
\graphicspath{ {./images/} }

\title{AgenticCADedit: A Stateful, Tool-Mediated Agentic Approach to Multimodal 3D CAD Editing}

\author{
Saptarshi Neil Sinha$^{*}$ \\
Fraunhofer IGD, Darmstadt, Germany \\
\texttt{saptarshineilsinha@gmail.com} \And
Mika Silvan Goschke \\
Fraunhofer IGD, Darmstadt, Germany \\
\texttt{mika.silvan.goschke@igd.fraunhofer.de} \And
Paul Julius K\"uhn \\
Fraunhofer IGD, Darmstadt, Germany \\
\texttt{julius.kuehn@igd.fraunhofer.de} \And
Arjan Kuijper\\
TU Darmstadt, Darmstadt, Germany \\
\texttt{arjan.kuijper@igd.fraunhofer.de} \And
Michael Weinmann \\
Delft University of Technology, Delft, Netherlands \\
\texttt{m.weinmann@tudelft.nl}
}

\begin{document}
\maketitle
% \renewcommand{\thefootnote}{\fnsymbol{footnote}}
% \footnotetext[1]{Equal contribution}
\renewcommand{\thefootnote}{\fnsymbol{footnote}}
\footnotetext[1]{Corresponding author.}

\input{sec/0_abstract}    
\input{sec/1_intro}
\input{sec/related_works}
\input{sec/methodology}
\input{sec/datasets}
\input{sec/evaluation}
\input{sec/conclusion}

\bibliographystyle{unsrt}  
\bibliography{main}  %%% Remove comment to use the external .bib 

\end{document}

%% file: sec/0_abstract.tex
\begin{abstract}
Computer-aided design is central to industrial manufacturing, and much of a designer's daily work
consists of editing existing models from multimodal requests involving speech, sketches,
and model interaction. Existing neural CAD approaches focus predominantly on unconditional or text-conditioned generation. The neuralCAD-Edit approach formalizes expert multimodal editing requests, but its iterative baseline refines a complete CAD program across attempts, executing each attempt from the original model in a stateless CAD environment. Every attempt must therefore reconstruct the entire edit from scratch, so partially correct progress is discarded rather than accumulated, and the model can neither inspect the geometry it has just produced nor selectively revert a single faulty operation. We present AgenticCADedit, which turns editing into a sequence of small, verifiable actions on a persistent CAD state instead of a single regenerated program. Rather than emitting one complete program, it applies incremental code steps that each commit to the session, inspects the resulting faces and edges, renders highlighted selections to verify that the intended region was addressed, and reverts individual operations when it was not. Subsequent actions therefore build on the geometry produced by earlier ones. Our approach improves on all metrics for all three evaluated LLMs (open-weight: qwen3.6-27b, gemma4-31b; proprietary: gpt-5.6-luna), with the largest gains for the weakest baseline model, qwen3.6-27b, whose validity rises from 51.0\% to 94.8\% and acceptance from 1.6\% to 12.0\%. A token-cost analysis with gpt-5.6-luna further shows $66.7$\% fewer output tokens than neuralCAD-Edit, while $94.8$\% of input tokens are served from the prompt cache.
\end{abstract}

%% file: sec/1_intro.tex
\section{Introduction}
% \label{sec:intro}
Computer-aided design (CAD) is the medium in which industrial products are specified. Automotive bodies, machine components, and electronic housings alike exist as CAD geometry before a single part is produced, and original equipment manufacturers (OEMs) therefore rely on it across the entire product lifecycle. The boundary representation (B-Rep)~\cite{weiler1986brep} stored by these systems remains the reference representation from the first concept to the finally released part. Yet design is rarely a single act of creation. Parts are refined to meet tolerances, corrected after review, adapted to supplier constraints, and updated whenever a specification changes. Editing an existing model therefore constitutes a substantial part of day-to-day design work.
Each such edit carries a communication cost. A reviewer describes a change, a designer interprets it, opens the model, locates the relevant faces and edges in the underlying mesh, performs the modification, and repeats the process whenever the intent was only partially understood. Since time to market is a decisive factor in industrial competition, the latency of this exchange is an economic quantity rather than merely a matter of convenience. Automating every part of this process can directly shorten design iterations. However, the difficulty is that engineers do not communicate design changes through text alone. They speak while rotating the model, point at the face they mean, sketch contours on top of the geometry, and refer back to previous actions, so the instruction is multimodal and interleaved in time~\cite{perrett2026neuralcadedit}. Moreover, editing is inherently cumulative: an operation changes the current geometry, subsequent operations act on the result, and an incorrect intermediate step may need to be inspected, revised, or reverted. A system for realistic CAD editing must therefore recover intent across modalities while interacting with an evolving model that it did not author.
\begin{figure*}[t]
\centering
\includegraphics[width=0.9\textwidth]{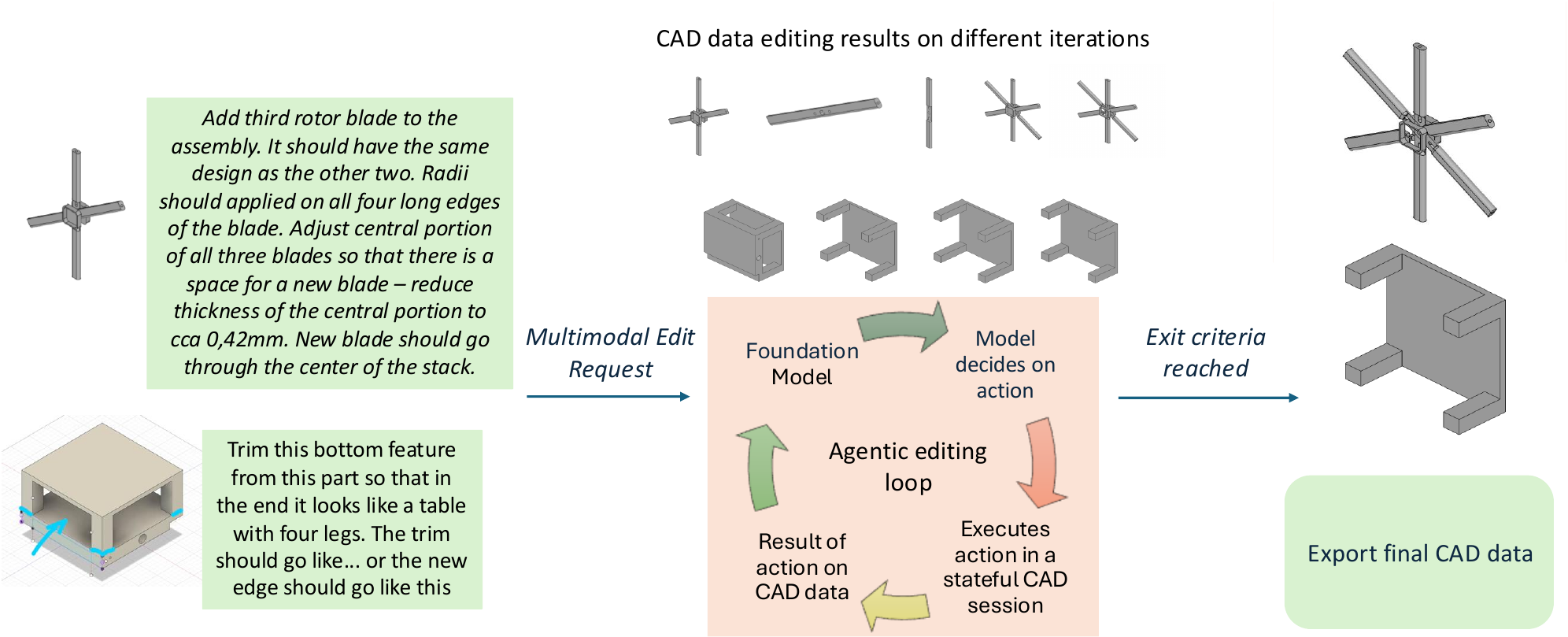}
\caption{\textbf{Overview of AgenticCADedit.} Given a multimodal edit request
and input CAD data, a foundation model incrementally edits a persistent CAD
state. It can execute code, inspect geometry, render highlighted entities, and
undo edits. Once the model declares completion, the result is exported.}
\label{fig:pipeline_agenticCADEdit}
\end{figure*}
Recent learning-based approaches have progressed from direct B-Rep synthesis~\cite{xu2024brepgen,jayaraman2022solidgen} to models that generate construction sequences~\cite{wu2021deepcad,khan2024text2cad} and, more recently, executable CAD scripts~\cite{rukhovich2025cadrecode,doris2026cadcoder}. Editing has received substantially less attention. Existing methods remain predominantly text-conditioned, either infilling command sequences~\cite{yuan2025cadeditor} or manipulating semantic latent representations~\cite{liu2025brepler}. neuralCAD-Edit~\cite{perrett2026neuralcadedit} is, to our knowledge, the only apprpach that moves beyond text for editing existing 3D solids, recording professional designers speaking, pointing, and drawing while manipulating a model.
The neuralCAD-Edit baseline provides feedback across attempts but no persistent editing state. Each iteration emits a complete CadQuery~\cite{cadquery} function executed from the original model and hence, its rendering and output inform the next attempt, but its geometric effects do not persist. Editing therefore corresponds to repeated \emph{full-program refinement}, i.e., the model reconstructs the complete edit from the original geometry instead of accumulating operations on an evolving model.
In a stateful CAD environment, the model under construction persists across operations, so each executed step updates it and later steps build on the resulting geometry instead of re-deriving it. Prior agentic systems have explored this setting, including CAD-Assistant~\cite{mallis2025cadassistant} and Embodied CAD~\cite{liu2026embodiedcad}. However, these systems address different problem settings and tool spaces. CAD-Assistant primarily targets tasks such as sketch reasoning, autoconstraining, and parameterization, while Embodied CAD constructs assemblies from structured specifications through a fixed skill library. Neither studies stateful interaction as the editing paradigm for general multimodal modifications of existing 3D solids under a reference-based editing benchmark. This leaves open whether persistent, tool-mediated interaction benefits multimodal CAD editing.
We investigate this question with \textbf{AgenticCADedit}, our stateful, tool-mediated approach to multimodal 3D CAD editing. To our knowledge, it is the first approach to combine video-recorded multimodal edit requests with stateful, tool-mediated editing of existing 3D CAD models. An LLM operates inside a persistent CadQuery~\cite{cadquery} session exposed through a Model Context Protocol (MCP) server, applying incremental code steps rather than regenerating the complete edit. It can inspect faces and edges numerically, render highlighted selections, and undo unsuccessful operations. These tools form a perception--decision--action--verification loop in which the model chooses whether to inspect, verify, edit, revise, or terminate. State-changing calls modify the same session, so later actions operate on geometry and variables produced by earlier ones. Furthermore, we evaluate AgenticCADedit on the benchmark provided by neuralCAD-Edit~\cite{perrett2026neuralcadedit} without modifying the benchmark or its evaluation protocol. This allows us to compare stateful incremental editing directly against the benchmark's iterative but stateless full-program refinement baseline across both open-weight and proprietary foundation models, while additionally measuring the token cost of the resulting sessions. Our contributions are:

\begin{itemize}
\item We introduce multimodal editing of existing 3D solids as \emph{stateful incremental interaction} rather than repeated full-program refinement. Each successful edit modifies a persistent CAD state on which subsequent actions operate, avoiding regeneration of the complete edit from the original model at every iteration.

\item We design an editing-oriented MCP tool interface around a persistent CadQuery session that combines incremental code execution with numerical geometry inspection, face and edge enumeration, rendered verification of selected entities, and explicit state management through undo operations.

\item We isolate the effect of this interaction paradigm on neuralCAD-Edit~\cite{perrett2026neuralcadedit}, keeping the benchmark and evaluation protocol fixed so that only the interaction paradigm differs from the stateless script-refinement baseline. Our approach improves all metrics for every evaluated open-weight and proprietary model while emitting fewer output tokens, with the largest gains for qwen3.6-27b.
\end{itemize}

%% file: sec/related_works.tex
\begin{table}[htb!]
\centering
\caption{MCP Tools available to the LLM during the agentic editing loop.}
\label{tab:tools}
\footnotesize
\setlength{\tabcolsep}{3pt}
\renewcommand{\arraystretch}{1.1}
\scalebox{0.9}{
\begin{tabular}{@{}l >{\raggedright\arraybackslash}p{0.31\linewidth} >{\raggedright\arraybackslash}p{0.35\linewidth}@{}}
\toprule
Tool & Purpose & Returns \\
\midrule
\multicolumn{3}{@{}l}{\textit{Edit}} \\
\texttt{run\_step} & Executes CadQuery code in the persistent session & \texttt{inspect} summary and program output, or error message with traceback \\
\addlinespace[3pt]
\multicolumn{3}{@{}l}{\textit{Query}} \\
\texttt{inspect} & Basic properties of the current model & Face and edge counts, axis-aligned bounding box  \\
\texttt{list\_faces} & Enumerates faces, sorted by area & Surface type, area, center, outward normal per face; total count \\
\texttt{list\_edges} & Enumerates edges, sorted by length & Edge type, length, center, radius for circles; total count \\
\addlinespace[3pt]
\multicolumn{3}{@{}l}{\textit{Render}} \\
\texttt{render\_views} & Renders selected viewpoints, optionally highlighting faces or edges & One image per requested view \\
\addlinespace[3pt]
\multicolumn{3}{@{}l}{\textit{State}} \\
\texttt{export\_cad} & Exports the current CAD  & Confirmation and path to the exported file \\
\texttt{undo} & Reverts the last successful edit & Updated model summary \\
\texttt{reset} & Clears the complete session and edit history, returning to an empty session & Confirmation that the session was reset \\
\addlinespace[3pt]
\multicolumn{3}{@{}l}{\textit{Virtual tool}} \\
\texttt{finish} & Signals that the LLM considers the edit complete & Terminates the agentic loop after the current batch of tool calls \\
\bottomrule
\end{tabular}
}
\end{table}
\section{Related works}
\label{sec:related_works}
\subsection{Neural CAD generation and representation}
In industrial domains such as manufacturing and mobility, 3D data is predominantly represented in CAD systems as a boundary representation (B-Rep)~\cite{weiler1986brep,lambourne2021brepnet}. A B-Rep describes a solid through surface geometry and explicit topology, exposing faces, edges, and vertices as addressable entities that support direct geometric editing. In contrast, voxels, point clouds, meshes, radiance fields, and splats~\cite{zhou2021pvd,luo2021dpm,chen2025sam3d,mildenhall2020nerf,kerbl2023splatting} represent shape through discrete samples, polygonal approximations, or learned fields and do not provide the same CAD-semantic structure for parametric editing. Consequently, methods developed for these representations generally do not produce editable B-Reps~\cite{li2025triposg,xiang2025trellis,parelli20253dlatte,li2025meshpad}, making CAD generation and editing a distinct research area.
\noindent\textbf{CAD benchmark datasets}
Several benchmark datasets have been created for this field to support learning-based approaches to
CAD generation and manipulation. At the largest scale, one million B-Reps are collected in
ABC~\cite{koch2019abc}, which has enabled substantial progress in geometric deep learning but keeps
only the final geometry and discards the construction history, while millions of
sketch-and-constraint graphs are provided by SketchGraphs~\cite{seff2020sketchgraphs} without
extending to 3D solid operations. Capturing design intent requires the construction timeline, which
only the Fusion 360 Gallery~\cite{willis2021fusion360,willis2022joinable} and
DeepCAD~\cite{wu2021deepcad} expose, the former with human-designed sequences at a small scale of
8.6k models and the latter with 178k sequences parsed from the public Onshape documents behind ABC.
Both remain restricted to sketch-and-extrude operations, and operation types and step order are
annotated in CC3D-Ops~\cite{dupont2022cadopsnet} with per-face labels rather than replayable
programs.
\noindent\textbf{Generative CAD approaches} In generative CAD two main approaches exist, and they differ in the representation they produce. The
first generates B-Reps directly, through
diffusion~\cite{wang2024vqcad,xu2024brepgen,lee2025brepdiff,liu2025hola}, autoregressive
decoding~\cite{jayaraman2022solidgen,xu2025autobrep,li2025brepgpt} and single-view
reconstruction~\cite{li2025caddreamer},
supported by geometry encoders such as UV-Net~\cite{jayaraman2021uvnet}, which yields valid solids but
no construction history and usually only a single body. The second models the construction sequence
itself, as a token
stream~\cite{wu2021deepcad,ganin2021cadlang,para2021sketchgen,nash2020polygen,willis2021sketchgen} or
through hierarchical codebooks~\cite{xu2022skexgen,xu2023hnc}, conditioned on point
clouds~\cite{khan2024cadsignet,dupont2024transcad,ma2023multicad}, or on text, images and structural
masks~\cite{khan2024text2cad,wang2025textcadvisual,wu2024cadvlm,chen2025cadcrafter,li2025cadllama,zhang2025flexcad,xu2024cadmllm,wang2025cadgpt}, with
Text2CAD~\cite{khan2024text2cad} showing that language can drive parametric construction while
inheriting DeepCAD's narrow operation vocabulary.  A more recent line of work sidesteps this fixed
command set by emitting general CAD scripts, as in CAD-Recode~\cite{rukhovich2025cadrecode}, which
reconstructs executable CadQuery~\cite{cadquery} code from point clouds and is trained on
procedurally generated sketch-extrude programs, CAD-Coder~\cite{doris2026cadcoder}, which generates
such code from images, and Text-to-CadQuery~\cite{xie2025text2cadquery}, which generates it from text
descriptions. We follow this script view
and operate on executable CadQuery code, since it covers the operations a real edit requires, such as
fillets, chamfers, shells and Booleans, and stays directly interpretable.
\subsection{Agentic CAD modelling}
Agentic AI for CAD modelling has become an active area of research, with a growing
body of work placing a foundation model in a loop with an executable CAD
environment rather than decoding a design in one pass. Some approaches write CAD
scripts, run them and repair them from the resulting execution errors~\cite{badagabettu2024query2cad,li2025cadllama,zuo2025cadhllm,yu2025gencad3d,ataei2026zerotocad,barkley2026cadsmith},
while others directly operate CAD-specific tools through vision-language models (VLMs)~\cite{mallis2025cadassistant,liu2026embodiedcad,gong2026toolcad,nithyanantham2025mcp4ifc}.
A common observation is that execution feedback alone is insufficient for geometric tasks, since
the reported failures are locally valid but globally incoherent, which has motivated rendered feedback
for iterative refinement~\cite{yuan2024premise,deng2025bimgent,wang2025textcadvisual,hu2026itercad} and VLM-based
raters for text-to-3D alignment~\cite{wu2024gpt4veval,khan2024text2cad}.
IterCAD~\cite{hu2026itercad} embeds an editing task in a multi-turn sandbox. Its edit 
instructions are text-only and its source programs are
synthetic, obtained by degrading correct scripts. Its multimodal input consists of the engineering
drawings that define the target part. Prior agentic CAD systems differ in how they carry information across turns. IterCAD~\cite{hu2026itercad} and the neuralCAD-Edit~\cite{perrett2026neuralcadedit} baseline re-provide the interaction history but re-execute a complete program from the original model in every attempt. CAD-Assistant~\cite{mallis2025cadassistant} and Embodied CAD~\cite{liu2026embodiedcad} maintain a persistent session, yet address different tasks, namely question answering, autoconstraining and sketch parameterization in the former case and assembly construction from structured specifications in the latter, without comparing the resulting geometry to a reference model. We apply persistent, tool-mediated interaction to multimodal edits of existing solids under a reference-based benchmark, where the agent modifies one CAD state incrementally, verifies intermediate geometry visually and can undo an unsuccessful operation, so later actions build on earlier edits. A second limitation is methodological. Each agent is compared against a weaker or differently trained
system, and several are fine-tuned for the
task~\cite{hu2026itercad,gong2026toolcad,liu2026embodiedcad}, so the contribution of the interface
cannot be separated from the contribution of the training. We take inspiration from these systems and
address multimodal edits of an existing CAD model, holding the model, the sampling parameters and the benchmark fixed so that only the interface varies. 
%Coordinate-based selection is a further known
% weakness of script-based editing, which LLM4CAD-Editor~\cite{sun2026llm4cadeditor} avoids through
% named entities in a domain-specific language and which our interface exposes through visual
% confirmation instead.
\subsection{Multimodal CAD editing and benchmarks}
CAD editing has been far less explored than generation, and the conditioning
signals that are available remain narrow. Text-based approaches either treat
editing as a locate-then-infill problem over command sequences, evaluated on a
synthetic benchmark of single components~\cite{yuan2025cadeditor}, modify a
semantic B-Rep latent space with an LLM~\cite{liu2025brepler}, or edit through a
domain-specific language whose named entities replace
coordinates~\cite{sun2026llm4cadeditor}. Related resources
outside CAD are similarly restricted, since they rely on text-only
requests~\cite{parelli20253dlatte}, on a rendered goal image that presupposes the
completed edit~\cite{gu2025blendergym}, or on synthetic sketch
conditioning~\cite{li2025meshpad}. Two benchmarks move beyond text.
mrCAD~\cite{mccarthy2025mrcad} collects refinement instructions that combine text and drawings, and
reports that state-of-the-art VLMs follow generation instructions considerably better than refinement
instructions, but it is restricted to a 2D CAD environment.
neuralCAD-Edit~\cite{perrett2026neuralcadedit} is, to our knowledge, the only approach that moves
beyond text for 3D solids, collecting requests in which professional designers speak, point and
draw while manipulating the model, together with reference edits for single bodies
and assemblies. In this setting the limits of stateless refinement become most visible when compared against our approach. We evaluate on this benchmark and reuse its evaluation protocol unchanged, so any difference follows from the interaction paradigm.

%% file: sec/methodology.tex
\section{Methodology}
We introduce an agentic approach for editing CAD data from multimodal edit
requests. Instead of asking a large language model (LLM) to produce one complete
CAD program per iteration, our agentic CAD editing arm, shown in
Fig.~\ref{fig:pipeline_agenticCADEdit}, gives the model direct and incremental
control over a CAD session through a Model Context Protocol (MCP) interface. The LLM manipulates geometry by tool calls which are carried out by a harness, i.e.\ the program that drives the agentic loop and mediates between the LLM and a persistent CAD session.
Our agentic CAD editing approach therefore becomes a sequence of small and verifiable actions. The
following subsections describe the context given to the LLM, the execution
environment, the agentic loop, the exposed tools, the model interface and the
exit criteria.
\subsection{LLM context}
% conversation_instruction, tool schemas, instruction, task info, preloaded model
Before the agentic loop starts, the LLM receives a prompt consisting of a conversation
instruction, the schemas of the tools exposed by the MCP server, and the task information.
The conversation instruction states that the LLM works inside a persistent session of a
parametric CAD scripting framework, CadQuery~\cite{cadquery} in our case, in which the
framework is already imported, the model to be edited is already loaded, variables persist
across steps, the evolving geometry is always bound to a single result variable, and the
task is solved through tool calls. It further advises the LLM to inspect and render the
model before editing, to keep code steps small, to visually confirm the relevant faces and
edges before operating on them, to verify the result after each modification, and to report
completion only once the model matches the request. Following neuralCADEdit~\cite{perrett2026neuralcadedit}, we support the same multimodal edit requests,
namely text passed directly as a string and videos in which the designer speaks while
pointing at, rotating and drawing on the model.
\subsection{Agentic CAD editing loop}
The main agentic loop (see Figure~\ref{fig:pipeline_agenticCADEdit}) drives the interaction between the LLM and the CAD session.
In each iteration, the LLM is called with the complete conversation history and the
available tool schemas. If the LLM requests one or more tools, the harness
executes them in the order they were requested and appends their results to the
conversation, so that the LLM observes the effect of its own actions in the next
iteration. Several tools can be requested in a single turn, which allows the LLM
to combine an edit with a subsequent inspection or rendering without spending an
additional iteration on it. Every tool call is bounded by a time limit, so that a
single long-running operation cannot stall the whole run. The loop continues with
the updated conversation until one of the exit criteria is met, at which point the
current CAD model is exported for evaluation. 

\subsubsection{Stateful execution environment}
The harness launches the MCP server as a subprocess and communicates over standard input and output. Its persistent namespace imports CadQuery once and retains LLM-defined variables across calls, allowing helper variables to be reused instead of regenerating a complete program. The current geometry is stored in \texttt{result}. Before every code execution the server pushes the session state onto a history
stack, so that a failed step restores the previous state and never leaves the session
inconsistent, while the \texttt{undo} tool reverts an edit that executed successfully but
produced an unintended result.
\begin{table*}[t]
\centering
\begin{minipage}[t]{0.49\textwidth}
\centering
\caption{Mean over all tasks ($N=192$). Failed or invalid edits are scored as zero. The LLM model is given in parentheses.}
\label{tab:mean_all_tasks}
\resizebox{\textwidth}{!}{%
\begin{tabular}{lrrrr}
\toprule
edit\_user & validity (\%) $\uparrow$ & instruction $\uparrow$ & quality $\uparrow$ & acceptance (\%) $\uparrow$ \\
\midrule
Human baseline & 100.00 & 0.75 & 0.75 & 82.30 \\
\midrule
neuralCAD-Edit~\cite{perrett2026neuralcadedit} (qwen3.6-27b) & 51.00 & 0.17 & 0.17 & 1.60 \\
\textbf{Ours} (qwen3.6-27b) & \textbf{94.80} & \textbf{0.32} & \textbf{0.39} & \textbf{12.00} \\
\midrule
neuralCAD-Edit~\cite{perrett2026neuralcadedit} (gemma4-31b) & 40.10 & 0.19 & 0.15 & 2.60 \\
\textbf{Ours} (gemma4-31b) & \textbf{57.30} & \textbf{0.23} & \textbf{0.21} & \textbf{6.80} \\
\midrule
neuralCAD-Edit~\cite{perrett2026neuralcadedit} (gpt-5.6-luna) & 96.40 & 0.57 & 0.57 & 50.00 \\
\textbf{Ours} (gpt-5.6-luna) & \textbf{99.00} & \textbf{0.59} & \textbf{0.60} & \textbf{53.60} \\
\bottomrule
\end{tabular}}
\end{minipage}
\hfill
\begin{minipage}[t]{0.49\textwidth}
\centering
\caption{Median over successful tasks. \texttt{n\_valid} is the number of tasks that produced a valid, gradable CAD output file. Valid subsets differ across rows; Tab.~\ref{tab:pairwise} provides the pairwise comparison.}
\label{tab:median_successful}
\resizebox{\textwidth}{!}{%
\begin{tabular}{lrrrr}
\toprule
edit\_user & n\_valid $\uparrow$ & chamfer\_dist $\downarrow$ & iou $\uparrow$ & DINO-sim $\uparrow$ \\
\midrule
Human baseline & 192.00 & 2.65 & 0.95 & 0.99 \\
\midrule
neuralCAD-Edit~\cite{perrett2026neuralcadedit} (qwen3.6-27b) & 98.00 & 4.26 & 0.64 & 0.53 \\
\textbf{Ours} (qwen3.6-27b) & \textbf{182.00} & \textbf{3.72} & \textbf{0.71} & \textbf{0.57} \\
\midrule
neuralCAD-Edit~\cite{perrett2026neuralcadedit} (gemma4-31b) & 77.00 & 9.41 & 0.54 & 0.62 \\
\textbf{Ours} (gemma4-31b) & \textbf{110.00} & \textbf{8.80} & \textbf{0.55} & \textbf{0.65} \\
\midrule
neuralCAD-Edit~\cite{perrett2026neuralcadedit} (gpt-5.6-luna) & 185.00 & 3.15 & 0.83 & 0.63 \\
\textbf{Ours} (gpt-5.6-luna) & \textbf{190.00} & \textbf{2.98} & \textbf{0.87} & \textbf{0.64} \\
\bottomrule
\end{tabular}}
\end{minipage}
\end{table*}
\begin{table*}[t]
\centering
\caption{
Pairwise comparison of mean scores. Both rows in a pair use the same LLM (given in parentheses), the same evaluation protocol, and are evaluated on the identical set of tasks, so any difference comes from the interaction paradigm alone.}
\label{tab:pairwise}
\resizebox{0.7\textwidth}{!}{%
\begin{tabular}{lrrrrrrr}
\toprule
edit\_user & chamfer dist $\downarrow$ & iou $\uparrow$ & Dino-sim $\uparrow$ & instruction $\uparrow$ & quality $\uparrow$ & acceptance (\%) $\uparrow$ & n\_tasks \\
\midrule
neuralCAD-Edit~\cite{perrett2026neuralcadedit} (qwen3.6-27b) & 66.60 & 0.56 & 0.59 & 0.18 & 0.34 & 3.20 & 95.00 \\
\textbf{Ours} (qwen3.6-27b) & \textbf{21.84} & \textbf{0.60} & \textbf{0.61} & \textbf{0.36} & \textbf{0.42} & \textbf{15.80} & 95.00 \\
\midrule
neuralCAD-Edit~\cite{perrett2026neuralcadedit} (gemma4-31b) & \textbf{58.47} & 0.49 & 0.60 & 0.23 & 0.35 & 5.60 & 71.00 \\
\textbf{Ours} (gemma4-31b) & 78.3 &\textbf{0.57} & \textbf{0.61} & \textbf{0.28} & \textbf{0.36} & \textbf{9.90} & 71.00 \\
\midrule
neuralCAD-Edit~\cite{perrett2026neuralcadedit} (gpt-5.6-luna) & 24.38 & 0.66 & 0.64 & 0.58 & 0.60 & 52.20 & 184.00 \\
\textbf{Ours} (gpt-5.6-luna) & \textbf{24.21} & \textbf{0.69} & \textbf{0.65} & \textbf{0.60} & \textbf{0.61} & \textbf{55.40} & 184.00\\
\bottomrule
\end{tabular}}
\end{table*}
\begin{figure}
    \centering
    \includegraphics[width=0.25\linewidth]{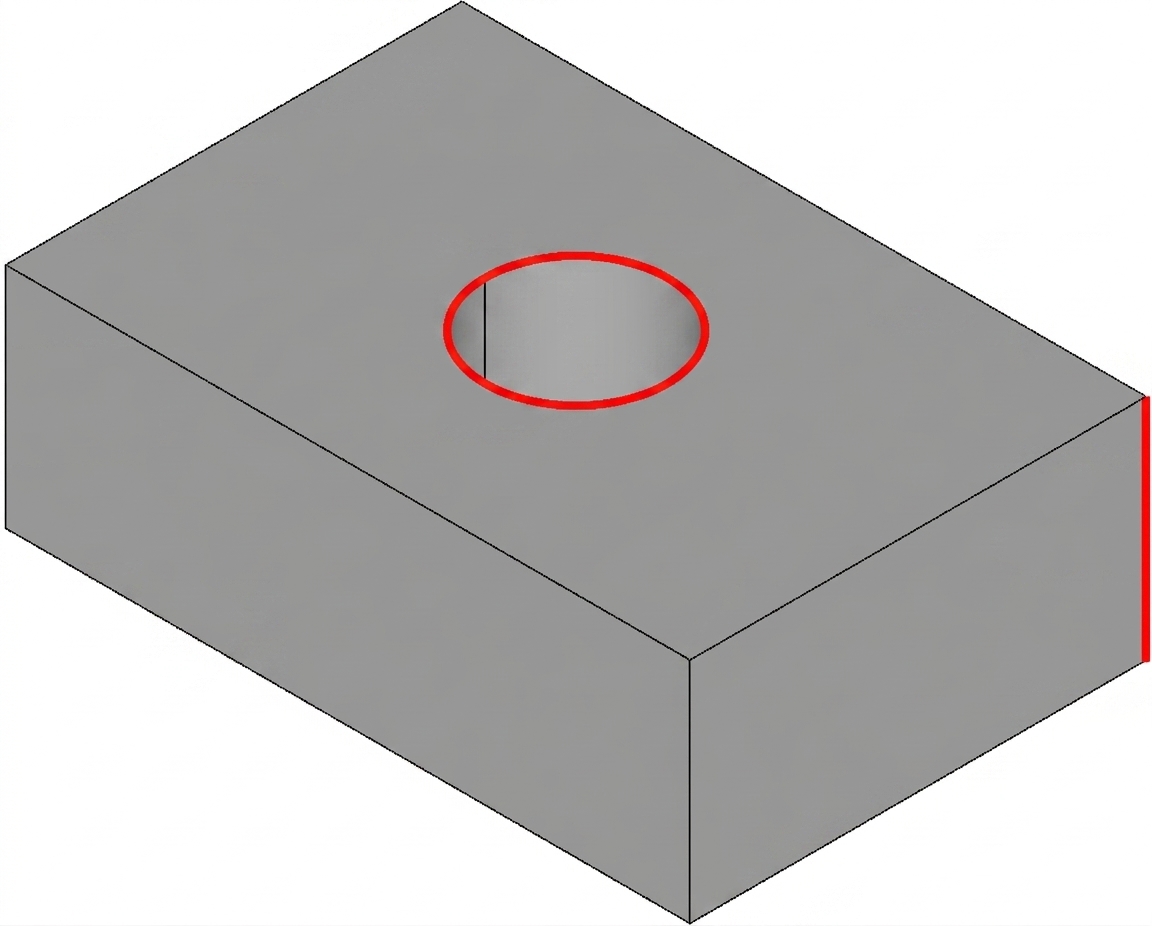}
    \caption{Two edges of the B-Rep selected, marked in red}
    \label{fig:RenderViews}
\end{figure}
\subsubsection{MCP tools}
The MCP tools define the action space of the LLM and they cover four kinds of
interaction with the session, namely executing code, querying the geometry
numerically, rendering it visually, and managing the session state. The numerical
tools are needed because the LLM cannot reliably infer dimensions or selector strings
from images alone, and the rendering tool is needed because numerical output does not
reveal whether the intended region was actually selected. We present the overview of
each tool in Table~\ref{tab:tools}. The core editing tool is \textbf{run\_step}, which executes CadQuery~\cite{cadquery} code
in the persistent session so that variables and progress carry over between calls, with
the current model always stored in \texttt{result} and the previous state restored on
error. The remaining tools let the LLM query the geometry numerically, verify visually
that a selector matched the intended region as shown in
Figure~\ref{fig:RenderViews}, manage the session state, and signal completion of the edit
through \textbf{finish}, which terminates the agentic loop.
\begin{figure*}[htb!]
    \centering
    \begin{minipage}[t]{0.48\textwidth}
        \centering
        \includegraphics[width=\textwidth]{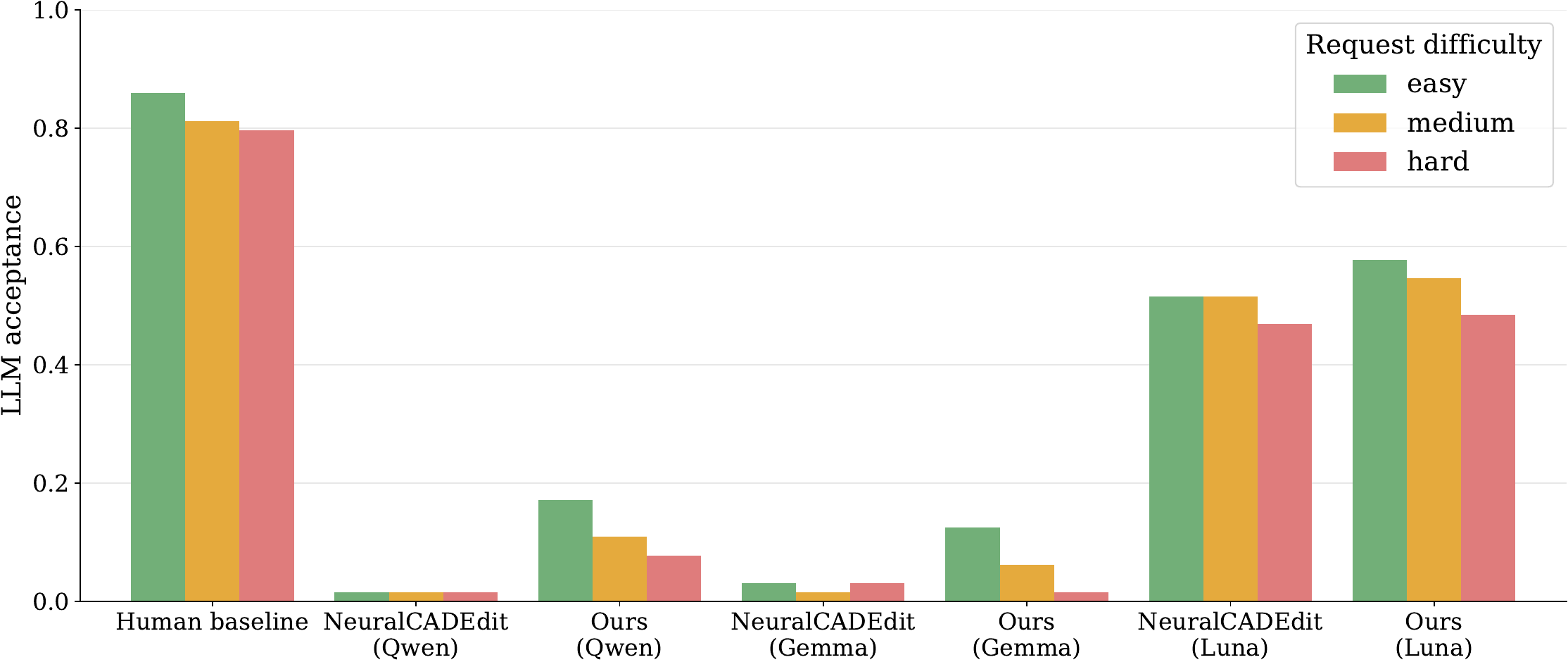}
        \caption{LLM acceptance grouped by request difficulty}
        \label{fig:AcceptanceByDifficulty}
    \end{minipage}
    \hfill
    \begin{minipage}[t]{0.48\textwidth}
        \centering
        \includegraphics[width=\textwidth]{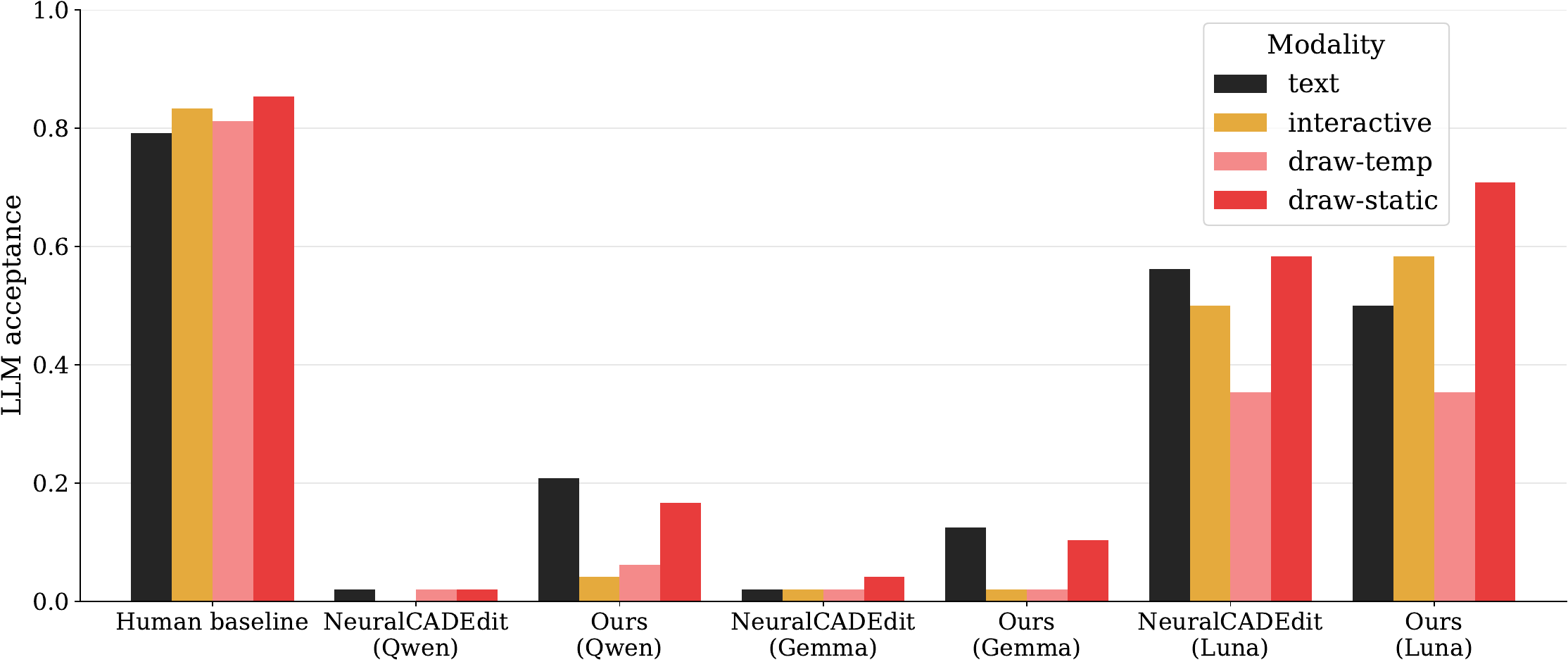}
        \caption{LLM acceptance grouped by input modality}
        \label{fig:AcceptanceByModality}
    \end{minipage}
\end{figure*}
\subsubsection{LLM interface}
We use the OpenAI tool-calling format so different models share one harness. Tool schemas accompany every request; model responses and textual tool results are appended unchanged with their call identifiers. Because tool messages cannot contain images, renders from a turn are collected in one additional user message. The harness records token counts, transcripts, and images, including for failed runs, enabling behavioral inspection and cost analysis.
\subsubsection{Exit criteria}
The loop ends in one of three ways. The intended case is that the LLM calls the
\texttt{finish} tool, which signals that it considers the edit complete.
The second case is that the LLM returns a response without any tool call, which is
treated as an implicit end of the trajectory. The third case is that the iteration
budget is exhausted, which bounds the cost of a run and prevents the LLM from
looping indefinitely on a step it cannot solve. Both approaches use a maximum of $100$ LLM iterations per task. In all three cases, the current
value of \texttt{result} is exported to a CAD format supported by our framework and passed on to the
evaluation, so that every run produces a model that can be compared against the
reference.

%% file: sec/datasets.tex
\section{Dataset}

We evaluate on the neuralCAD-Edit benchmark~\cite{perrett2026neuralcadedit}, which provides 192 expert multimodal CAD editing requests. The requests span four input modalities: \emph{text}, consisting of typed instructions without speech or direct model interaction; \emph{interactive}, in which designers speak while pointing at and manipulating the CAD model; \emph{draw-temp}, which additionally includes temporary drawings that disappear after a short period; and \emph{draw-static}, in which drawings remain visible throughout the request or until manually removed. The multimodal requests are recorded as videos, preserving the temporal alignment between speech, model interaction, viewpoint changes, and drawings. The benchmark further covers three difficulty levels and four model categories formed by the combinations of parametric versus non-parametric models and single-body versus assembly construction.

\section{Implementation details}
The open-weight models qwen3.6-27b~\cite{qwen3_6_27b_2026} and gemma4-31b~\cite{gemma4_31b_2026} were self-hosted on a GPU cluster equipped with four NVIDIA A100 GPUs (40\,GB each). In contrast, gpt-5.6-luna~\cite{gpt5_6_luna_2026} was accessed through its public API with the reasoning effort set to high.

%% file: sec/evaluation.tex
\section{Evaluation}
We follow the baseline metrics: Chamfer distance, voxel IoU, DINOv2 similarity on isometric renders~\cite{oquab2024dinov2learningrobustvisual}, validity, and VLM ratings for instruction following, model quality, and acceptance. Instruction and quality are rated from 1 to 7 and normalized to $[0,1]$; acceptance requires both scores to be at least 5. Opus 5 medium, operated inside Claude Code, produces all ratings by inspecting and comparing rendered regions. Unlike the earlier VLM evaluators considered in neuralCAD-Edit ~\cite{perrett2026neuralcadedit}, Opus 5 inside Claude Code closely reproduces the expert ratings on the human baseline, providing stronger evidence that its automatic ratings are meaningful for this benchmark. Our qualitative analysis uses renders and transcripts from easy, medium, and hard examples.

\subsection{Comparison with Baseline Approaches}
As the primary baseline for comparison, we use the iterative script-refinement baseline released with neuralCAD-Edit~\cite{perrett2026neuralcadedit}. We evaluate the same multimodal requests with and without the stateful session for proprietary gpt-5.6-luna~\cite{gpt5_6_luna_2026} and open-weight gemma4-31b~\cite{gemma4_31b_2026} and qwen3.6-27b~\cite{qwen3_6_27b_2026}. Using each model in both harnesses separates the interaction paradigm from the underlying model.

\begin{figure*}[htb!]
\centering
\includegraphics[width=0.9\textwidth]{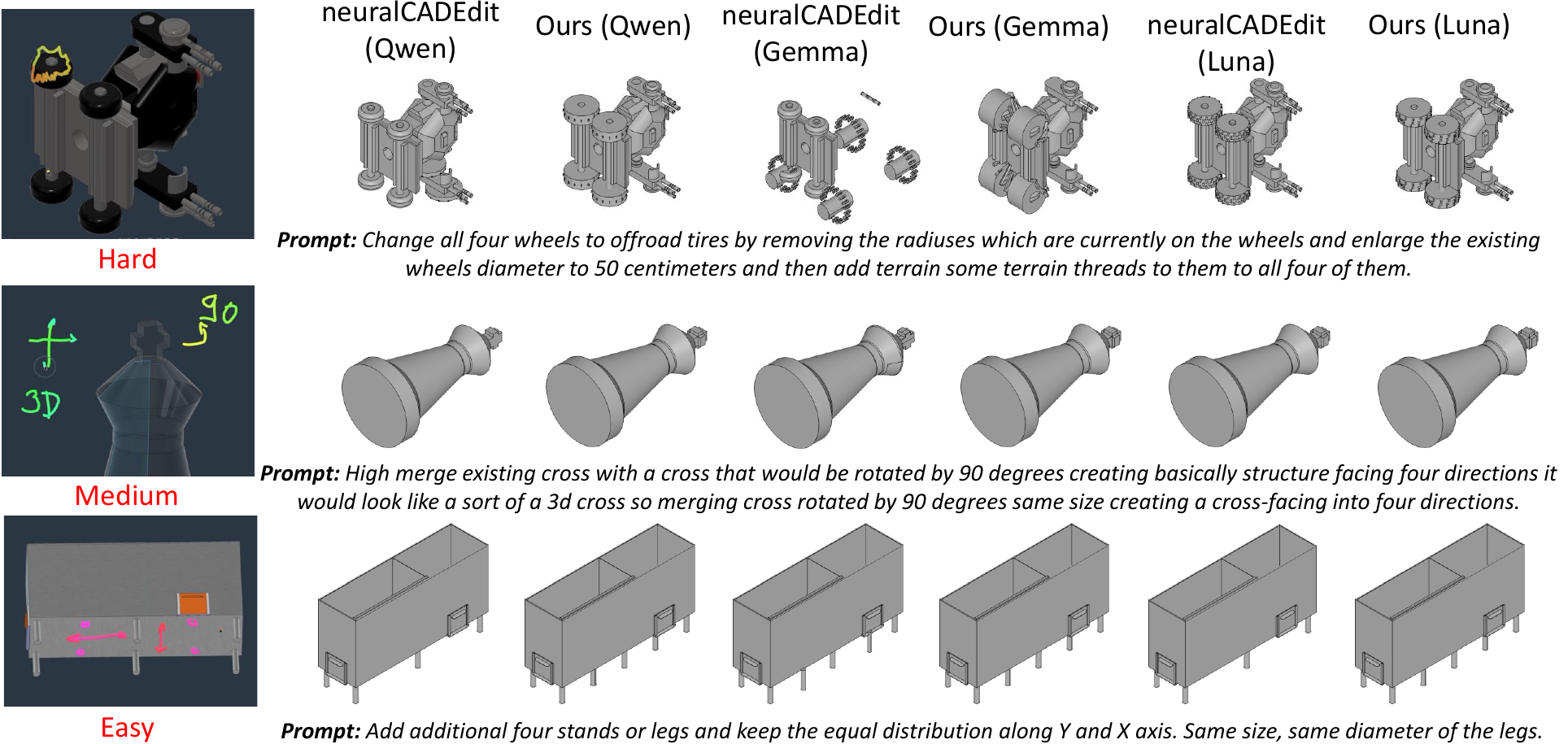}
\caption{Qualitative comparison across request difficulty. Each row shows one editing
request of easy, medium or hard difficulty, and the columns show the resulting edits of
neuralCAD-Edit~\cite{perrett2026neuralcadedit} and of our approach, each instantiated with
qwen3.6-27b~\cite{qwen3_6_27b_2026}, gemma4-31b~\cite{gemma4_31b_2026} and gpt-5.6-luna~\cite{gpt5_6_luna_2026}.}
\label{fig:qualitative_difficulty}
\end{figure*}

\subsection{Quantitative results}
Our approach improves over neuralCAD-Edit~\cite{perrett2026neuralcadedit} for every LLM and
on every metric of the mean evaluation over all $192$ tasks in Table~\ref{tab:mean_all_tasks},
where failed or invalid edits are scored as zero. The relative gains are largest for the
lightweight open weight models, since validity almost doubles and acceptance increases by more
than a factor of seven for qwen3.6-27b~\cite{qwen3_6_27b_2026}, suggesting that the more limited
open weight models benefit particularly strongly from the structured interaction provided by the
agentic session, whereas for gpt-5.6-luna~\cite{gpt5_6_luna_2026} the baseline is already strong
and the margins are smaller.

Since using only mean conflates how often a system produces usable output with how good that
output is, we also report (see Table~\ref{tab:median_successful}) the number of valid outputs together with
the median geometric scores over exactly those tasks. The number of valid outputs grows
significantly for the open weight models, while the median Chamfer distance, IoU and DINO
similarity improve by smaller margins, which suggests that our approach mainly converts
previously broken attempts into usable edits and that the accuracy of an already usable edit
depends more on the underlying model. Since these medians are computed over a different
subset for each row, we additionally restrict (see Table~\ref{tab:pairwise}) each pair to the tasks
that both systems solved validly and holds model, iteration budget and protocol fixed, so any
remaining difference comes from the interaction paradigm alone. Similar conclusions hold under
this stricter setting, and the only exception is the mean Chamfer distance, which is sensitive
to a few strongly displaced outputs and can therefore worsen even when all other metrics
improve. Across all three views the human baseline stays ahead, in particular on acceptance,
so there is clear scope for improvement.
We also evaluate acceptance in a more fine grained way.
Fig.~\ref{fig:AcceptanceByDifficulty} groups the acceptance rate by request difficulty into
easy, medium and hard requests, and Fig.~\ref{fig:AcceptanceByModality} groups it by input
modality into text, interactive, draw-temp and draw-static requests, in both cases for the
human baseline and for all three pairs of systems. Acceptance generally decreases with request
difficulty, most clearly for the human baseline and our agentic variants. Our approach yields
substantial gains for qwen3.6-27b~\cite{qwen3_6_27b_2026} and
gpt-5.6-luna~\cite{gpt5_6_luna_2026} across all difficulty levels and for
gemma4-31b~\cite{gemma4_31b_2026} on easy and medium requests, while hard requests remain
particularly challenging for the open weight models.
Interactive and draw-temp requests are particularly difficult for open-weight models, probably because cursor motion and transient annotations must be recovered from frame sequences. gpt-5.6-luna handles interaction substantially better, but draw-temp remains its weakest modality. Persistent draw-static annotations reduce this ambiguity and yield higher acceptance, especially for gpt-5.6-luna.
\begin{figure*}[htb!]
\centering
\includegraphics[width=0.85\textwidth]{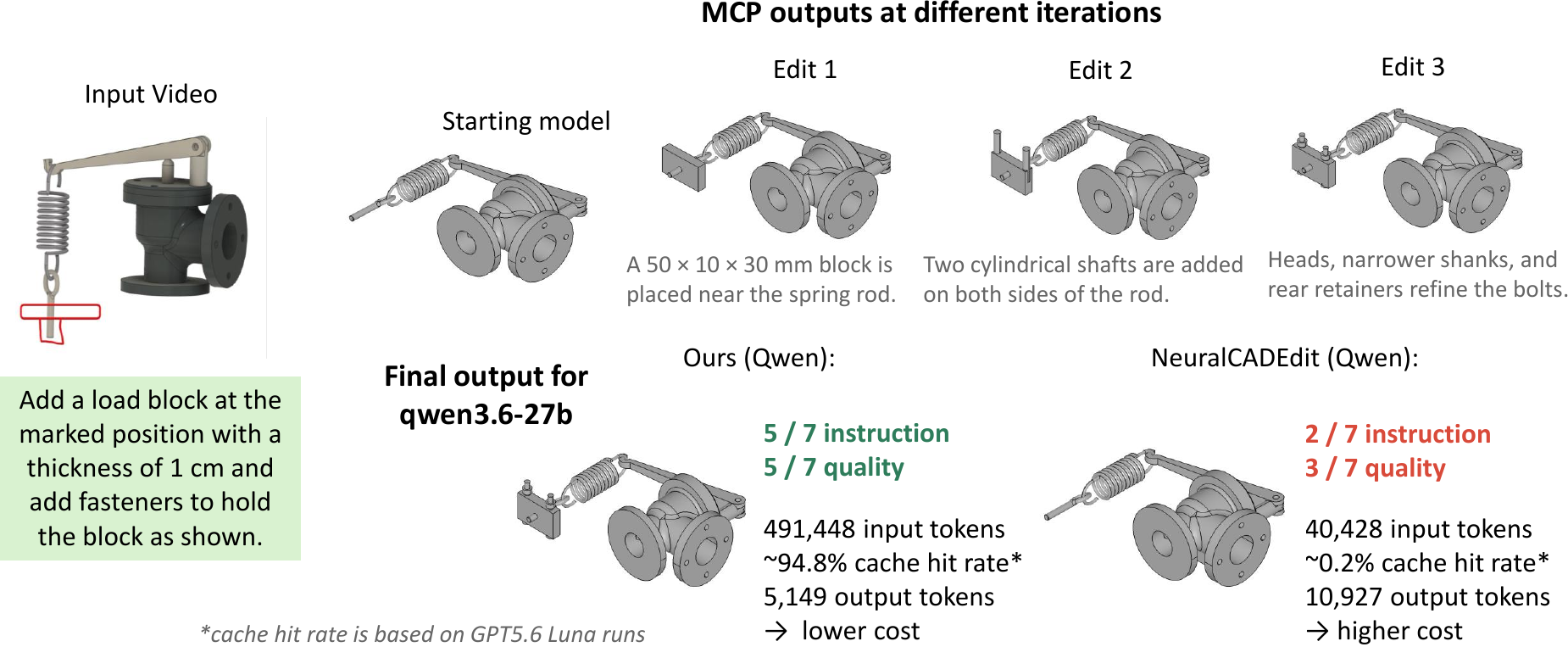}
\caption{Ablation of the agentic session. Given the starting model and the multimodal edit
instruction on the left, we show the intermediate outputs of our stateful agentic loop after
each operation, together with the final output of the neuralCAD-Edit
full-program refinement baseline~\cite{perrett2026neuralcadedit}. Both systems use
qwen3.6-27b~\cite{qwen3_6_27b_2026}, so the difference isolates the effect of operating
incrementally on a persistent CAD state instead of regenerating a complete edit program
across refinement attempts.}
\label{fig:qualitative_stages}
\end{figure*}
\subsection{Qualitative results}
We now qualitatively analyze the edits in Fig.~\ref{fig:qualitative_difficulty} to complement the previous quantitative evaluation with visual results. For the easy request, which asks for four
additional legs with equal distribution along the X and Y axes and the same size and diameter,
our approach preserves both the leg dimensions and the symmetry of the placement for all three
models. With qwen3.6-27b~\cite{qwen3_6_27b_2026} the baseline either changes the diameter of the added legs or places them
off the intended grid, and with gemma4-31b~\cite{gemma4_31b_2026} it frequently fails to produce a valid body at all,
whereas our approach recovers a usable and consistent result in both cases. With gpt-5.6-luna~\cite{gpt5_6_luna_2026} both
systems solve the request, so the difference reduces to the accuracy of the placement.

For the medium request, which asks to merge the existing cross with a copy rotated by ninety
degrees so that the result faces four directions, the gap between the models becomes more visible.
With qwen3.6-27b~\cite{qwen3_6_27b_2026} the baseline typically leaves the original cross almost unchanged or adds a
misaligned arm, while our approach produces the intended three dimensional cross. With gemma4-31b~\cite{gemma4_31b_2026}
the rotation is often applied to the wrong axis or only a partially merged body is created, and
even our approach does not always close the symmetry completely, whereas with gpt-5.6-luna~\cite{gpt5_6_luna_2026} the
baseline occasionally duplicates the geometry without merging it and our approach yields a clean
single body. The hard request combines removing wheel radiuses, enlarging wheel diameter, and adding treads to all four wheels. Open-weight models usually complete only the enlargement, while gpt-5.6-luna addresses more steps but leaves coarse treads. This mirrors the difficulty trend: the agentic session makes well-specified edits more reliable, while long multi-step requests remain challenging.

\subsection{Ablation studies}
The primary ablation of the agentic session is already contained in
Table~\ref{tab:mean_all_tasks}, Table~\ref{tab:median_successful} and
Table~\ref{tab:pairwise}. Both approaches use the same underlying LLM, but differ in how the CAD environment is exposed: the
neuralCAD-Edit~\cite{perrett2026neuralcadedit} baseline repeatedly refines a complete program
that is executed in a stateless CAD environment, whereas our approach performs incremental
tool-mediated edits on a persistent CAD state. The consistent improvements therefore quantify
the contribution of the interaction paradigm.

Fig.~\ref{fig:qualitative_stages} compares the final output of the full-program refinement
baseline with the intermediate states produced by our agentic loop for an instruction to add a
\(1\,\mathrm{cm}\)-thick load block at the marked location and secure it with fasteners. The neuralCAD-Edit~\cite{perrett2026neuralcadedit} baseline places an incorrectly sized and positioned block and produces incomplete fasteners. In
contrast, our loop first adds the load block, then introduces the bolt shafts, and finally refines
them with heads, shanks, and rear retainers. Each intermediate edit remains valid, and the final
result better follows the requested geometry and attachment structure. We additionally apply the
LLM-based instruction, quality and acceptance ratings described above to the example in
Fig.~\ref{fig:qualitative_stages}. Both systems use
qwen3.6-27b~\cite{qwen3_6_27b_2026}, so the comparison isolates the effect of the interaction
paradigm. Our final output receives an instruction score of 5 out of 7 and a quality score of
5 out of 7, while the full-program refinement baseline reaches only 2 out of 7 and 3 out of 7.

\subsection{Token cost}
Across all 192 tasks with gpt-5.6-luna~\cite{gpt5_6_luna_2026}, our loop processes
301.37 million input tokens versus 99.39 million for
neuralCAD-Edit~\cite{perrett2026neuralcadedit}, but 94.8\% are served from the prompt cache
compared with only 0.2\% for the baseline. This high cache reuse follows from the append-only
multi-turn session, whose successive requests share a largely unchanged prefix. Our loop also
reduces output tokens from 5.39 million to 1.80 million (66.7\%). Consequently, the recorded
API cost is only \$11.00 for our loop versus \$26.30 for the baseline, a reduction of 58.2\%,
despite the higher nominal input-token volume.

%% file: sec/conclusion.tex
\section{Conclusion}
We presented AgenticCADedit, an approach that reformulates multimodal CAD editing from
stateless full-program refinement into stateful, tool-mediated interaction with an evolving CAD
model. On the neuralCAD-Edit benchmark with $192$ expert multimodal requests, our approach
increases validity, instruction following, model quality and acceptance for every evaluated LLM,
with the largest gains for open-weight models, where validity almost
doubles and acceptance increases by more than a factor of seven for qwen3.6-27b. The pairwise
comparison on identical task sets, with model, iteration budget and evaluation protocol held fixed,
further isolates the effect of the interaction paradigm from that of the underlying foundation
model. Rather than repeatedly reconstructing a complete edit from the original geometry, the model
can act on, inspect, verify and revise a persistent geometric state.
Our gpt-5.6-luna cost analysis further shows $66.7$\% fewer output tokens, while $94.8$\% of input tokens are served from the prompt cache. 

The human baseline nevertheless remains ahead, and long composite requests and sketch-based intent remain only partially solved. To address this limitation, future work should introduce an explicit planning stage over the
requested operations and tighter grounding of annotated video input to the corresponding geometric
entities. Beyond this, a richer set of query and verification tools could further reduce the
remaining gap to expert edits.